\documentclass{article}
\usepackage{arxiv}
\usepackage[utf8]{inputenc} 
\usepackage[T1]{fontenc}    
\usepackage{hyperref}       
\usepackage{url}            
\usepackage{booktabs}       
\usepackage{amsfonts}       
\usepackage{nicefrac}       
\usepackage{microtype}      
\usepackage{lipsum}
\usepackage{graphicx}
\usepackage{comment}
\usepackage{multirow}
\usepackage{adjustbox}
\usepackage{subcaption}
\usepackage{amsmath}
\usepackage[psamsfonts]{amssymb}
\usepackage{amsxtra}
\usepackage{threeparttable}

\title{Statistical Attention Localization (SAL): Methodology
and Application to Object Classification}

\author{
  Yijing Yang, Vasileios Magoulianitis, Xinyu Wang and C.-C. Jay Kuo \\
  University of Southern California, Los Angeles, California, USA \\
  \texttt{yijingya@usc.edu, magoulia@usc.edu, xwang350@usc.edu, cckuo@sipi.usc.edu}
}

\begin{document}
\maketitle

\begin{abstract}

A statistical attention localization (SAL) method is proposed to facilitate the object classification task in this work. SAL consists of three steps: 1) preliminary attention window selection via decision statistics, 2) attention map refinement, and 3) rectangular attention region finalization. SAL computes soft-decision scores of local squared windows and uses them to identify salient regions in Step 1. To accommodate object of various sizes and shapes, SAL refines the preliminary result and obtain an attention map of more flexible shape in Step 2. Finally, SAL yields a rectangular attention region using the refined attention map and bounding box regularization in Step 3.  As an application, we adopt E-PixelHop, which is an object classification solution based on successive subspace learning (SSL), as the baseline. We apply SAL so as to obtain a cropped-out and resized attention region as an alternative input. Classification results of the whole image as well as the attention region are ensembled to achieve the highest classification accuracy. Experiments on the CIFAR-10 dataset are given to demonstrate the advantage of the SAL-assisted object classification method. 

\end{abstract}

\section{Introduction}\label{sec:introduction}

Given images containing both foreground objects and background, humans
pay more attention to object regions automatically for object
recognition. In computer vision, attention localization attempts to
mimics the human vision system in selecting the foreground object to
facilitate the recognition task. Most state-of-the-art attention
localization algorithms use deep-learning (DL) networks trained by
backpropagation. In this work, we propose an alternative attention
localization method by exploiting the soft decision made on a local
window based on the statistical principle. 

Our high-level idea can be best explained with a simple example. Consider
the classification of dog and cat images using the machine learning
technique. Each image has only one label to denote the main object in
the image. By partitioning input images into smaller local squared
windows, we can group these windows into clusters based on visual
similarity (i.e., the Euclidean distance between ordered pixels).  Some
clusters contain the image background such as the floor, wall, window,
furniture, etc., which are shared by images labeled by dog or cat. These
clusters have a higher entropy value against sample labels. In contrast,
other clusters contain regions more specific to a particular object
class, say, dog/cat facial regions (e.g., mouths, ears and eyes). Then,
samples in these clusters will be dominated by one class and they have a
lower entropy value sample labels. The resulting scheme is called
statistical attention localization (SAL). We will present a way
to implement SAL in this method and show its effectiveness in the
object classification task.

The proposed SAL method consists of three steps: 1) preliminary
attention selection, 2) attention map refinement, and 3) attention
region finalization.  In Step 1, SAL computes soft-decision scores of
local squared windows and uses them to identify salient regions. To
accommodate objects of various sizes and shapes, SAL refines the
preliminary result and obtain an attention map in Step 2. Finally, SAL
uses the refined attention map and bounding box regularization to yield
a rectangular attention region in Step 3.  As an application, we adopt a
state-of-the-art object classification solution based on successive
subspace learning (SSL), called E-PixelHop \cite{yang2021pixelhop}, as
the baseline. We apply SAL to obtain a cropped-out and resized
attention region as an alternative input. Classification results of the
whole image as well as the attention region are ensembled to achieve the
highest classification accuracy.  Experiments on the CIFAR-10 dataset
are given to demonstrate the advantage of the SAL-assisted object
classification method. 

The rest of the paper is organized as follows. Related work is reviewed
in Sec. \ref{sec:review}. The SAL method is presented in
Sec.~\ref{sec:attention}.  Experimental results are shown in
Sec.~\ref{sec:experiments}. Finally, concluding remarks are given in
Sec.~\ref{sec:conclusion}. 

\section{Review of Related Work}\label{sec:review}

{\bf Attention Localization and Learning.} Research has been done in
understanding decisions made by an object recognition system by
visualizing captured attention regions \cite{zeiler2014visualizing,
zhou2018interpretable, zhang2018interpretable}.  Since the soft decision
function is differentiable, attention can be learned by neural networks
through end-to-end optimization.  Global average pooling was used in
\cite{zhou2016learning} to localize discriminative regions learned by
convolutional neural networks (CNNs). The progressive attention
detection algorithm \cite{seo2016progressive} suppresses irrelevant
regions in the input image gradually and uses contexts of each local
patch to estimate an increasingly finetuned attention map.  

{\bf Attention-Assisted Object Classification.} Attention or object
localization has been exploited to improve the classification
performance furthermore, e.g.,~\cite{jetley2018learn, lin2015deep,
wang2019sharpen, zhou2016learning, zeiler2014visualizing, lin2015deep}.
It is observed that the classification performance can be improved by
focusing on the most important regions. The relationship between
attention estimated by local features and classification made by global
ones was analyzed in ~\cite{jetley2018learn} using a compatibility score
function. Class-specific attention was used to resolve confusion between
similar classes in \cite{wang2019sharpen}.

{\bf Attention in Fine-Grained and/or Small Object Classification.}
Research on object or part localization is widely adopted by
fine-grained and/or small object classification. Since labels are
usually available in the image level, human annotation on parts of the
object under a strong supervision assumption was considered in
\cite{zhang2014part,huang2016part}. Yet, the annotation task is
laborious and expensive. Besides, annotated key points or boxes could be
biased between individuals.  Annotation can be achieved by including
humans in the loop; namely, humans click on or mark discriminative
regions, or answer questions for visual recognition as introduced in
~\cite{branson2010visual, deng2013fine, duan2012discovering}.

{\bf Weakly-Supervised Attention Localization.} The object localization
problem can be solved by weakly supervised learning
\cite{wang2014weakly, crandall2006weakly, oquab2015object,
cinbis2016weakly,bazzani2016self}. Discriminative parts were localized
by training positive and negative image patches in
\cite{zhang2016picking}.  A self-taught object localization method was
proposed in ~\cite{bazzani2016self}. It identifies object regions by
analyzing the change in recognition scores when masking out different
parts of the input image. 

{\bf SSL-based Object Classification.} The successive subspace learning
(SSL) methodology was recently proposed by Kuo {\em et al.} in a
sequence of papers \cite{kuo2016understanding, kuo2017cnn, kuo2018data,
kuo2019interpretable}. SSL-based methods learn feature representations
in an unsupervised feedforward manner based on multi-stage principal
component analysis (PCA). Joint spatial-spectral representations are
obtained at different scales, where each scale is called a \textit{hop}. Based on the SSL framework, a few object
classification solutions have been developed. Examples include PixelHop
\cite{chen2020pixelhop}, PixelHop++ \cite{chen2020pixelhop++} and
E-PixelHop~\cite{yang2021pixelhop} methods.  They follow the traditional
pattern recognition paradigm by decomposing a classification problem
into two parts - the feature extraction and the decision making. 

\section{Statistical Attention Localization (SAL)}\label{sec:attention}

The SAL method is presented in this section. It localizes the attention
region using the statistical principle.  It consists of three steps: 1)
preliminary attention window selection, 2) attention map refinement,
and 3) attention region finalization. We use the classification problem
on the CIFAR-10~\cite{cifar10} dataset as an illustrative example. 

\subsection{Step 1: Preliminary Attention Window Selection}
\label{sec:pre_attention_window}

In the first step, SAL computes soft-decision scores of local squared
windows by solving a pixel-wise classification problem. It consists of
the following three substeps:
\begin{itemize}
\item[a] Context vector extraction: \\
Joint spatial-spectral features at various scales are extracted based on
the linear combination of neighbor pixels through 9 cascaded PixelHop++
units \cite{chen2020pixelhop++} (see Fig.~\ref{fig:feat_att_1}).  No
pooling is used so that extracted features can be easily aligned
spatially between different hops.  We obtain 9 filter response sets from
9 hop units at each pixel.  Each set corresponds to a growing receptive
field from shallow to deep hops. The concatenated response vector forms
a feature pyramid at each location.  It is called the \textit{context
vector} as shown in Fig.~\ref{fig:feat_att_1}. The receptive field in
the deepest hop has a size of 19x19. 

\begin{figure}[t]
\begin{center}
\includegraphics[width=0.9\linewidth]{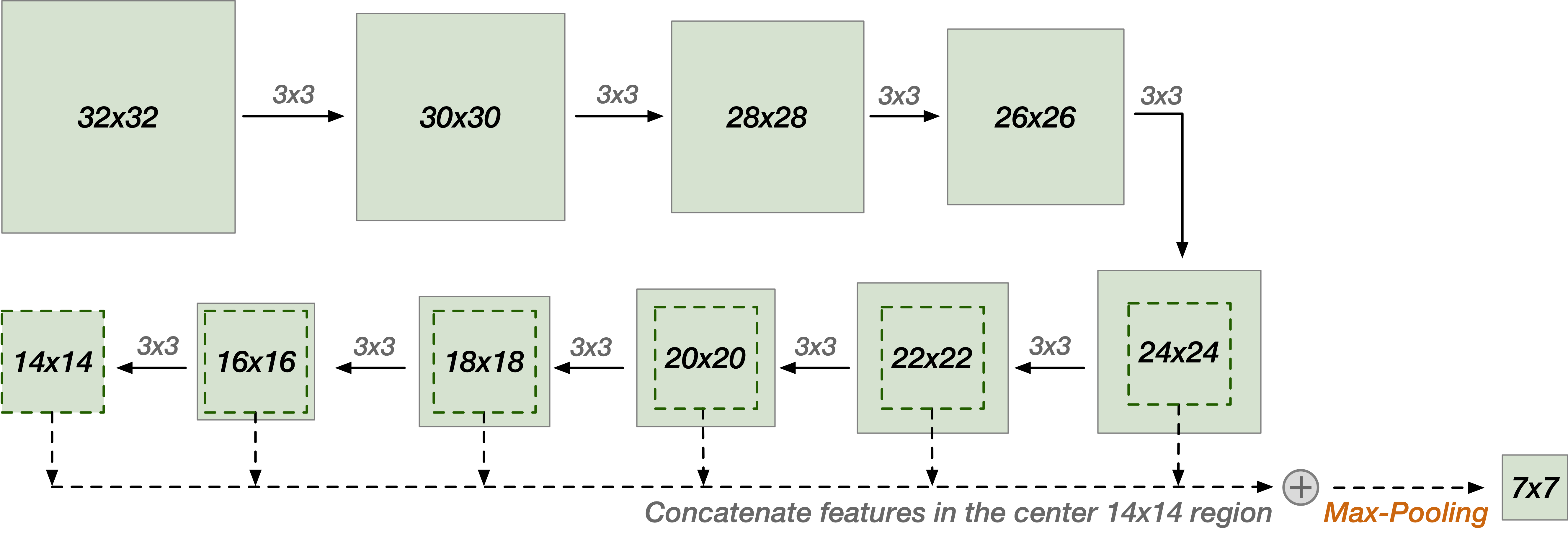}
\end{center}
\vspace{-10pt}
\caption{Illustration of the context vector extraction procedure, where
only the spatial dimension is shown.}\label{fig:feat_att_1}
\end{figure}

\item[b] Attention window selection: \\
After obtaining the context vector for each pixel, pixel-level
classification is conducted to yield soft-decision scores, which will be
used to estimate the discriminant power of each pixel.  To reduce the
complexity, we do a (2x2)-to-(1x1) max-pooling to generate 7x7 sparser
pixels in an image. Image labels are used as pixel labels in the
training phase. In the inference phase, the predicted soft decision is
in form of a tensor of size 7x7x10, where 10 is the class number in
CIFAR-10.  To get aligned with the spatial resolution in the deepest hop
unit, bilinear interpolation is applied to the soft decision channel-wise in order to
increase its spatial resolution by two.  Then, we can
compute the confidence level for each pixel in the 14x14 region. The
confidence level is defined as the inverse probability of most probable class based on the soft decision
vector at the pixel.  The higher the confidence level, the pixel is more discriminant for the top class in the soft
decision. The most discriminant pixel from the 14x14 region is selected
as the center of the attention window and its neighborhood of size WxW
is set as the preliminary attention window, where W is an odd number. 
\end{itemize}

The hyper-parameter, $W$, is set to 19 for the CIFAR-10 dataset in this
work, which is equal to the size of the receptive field in the deepest
hop.  Note also that the spatial resolution of CIFAR-10 images is 32x32,
which is already small and cropped for the object. On one hand, if the
attention window size is larger than 19x19, it would be very close to
the original input. On the other hand, if the attention window size is
significantly smaller than 19x19, it would not be stable and/or reliable. 

\subsection{Step 2: Attention Map Refinement}\label{sec:attention_refine}

\begin{figure}[t]
\begin{center}
\includegraphics[width=0.9\linewidth]{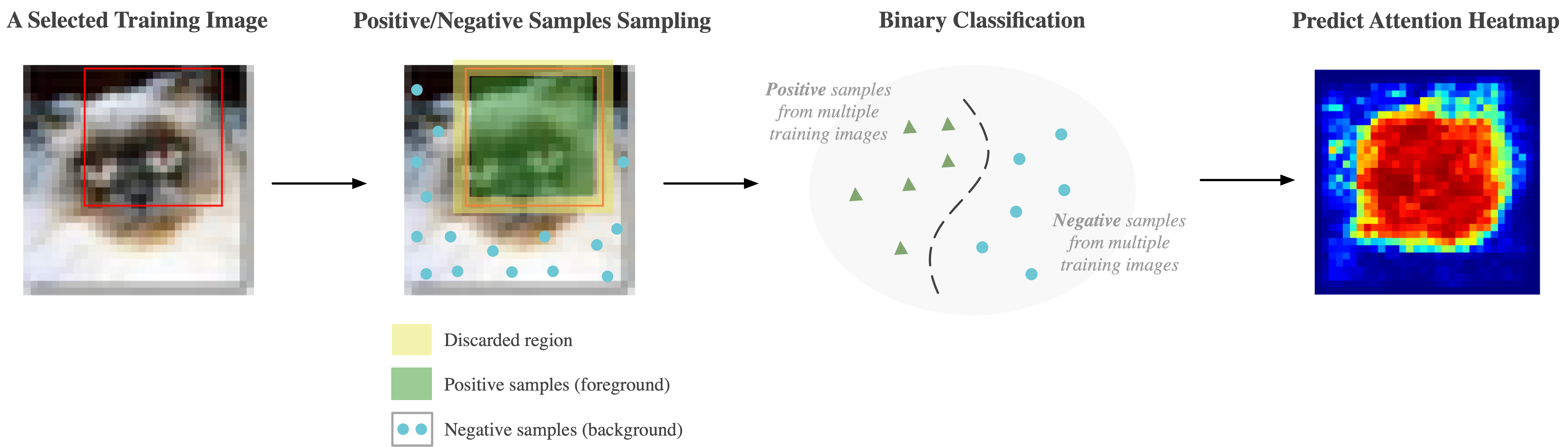}
\end{center}
\caption{Illustration of the attention map refinement step.}
\label{fig:pipe_attWeaklySuper}
\end{figure}
Although the attention region of an input can be well localized in Step
1, the 19x19 squared window is not ideal for all images due to the
variation of object sizes, shapes and orientations.  We adopt a weakly
supervised scheme to refine the attention map as illustrated in
Fig.~\ref{fig:pipe_attWeaklySuper}. It is formulated as a binary
classification problem between the positive class (i.e., attention
or foreground pixels) and the negative class (i.e., background pixels).
The refined attention map is the area with a higher positive class
probability. 

Given a selected subset of training images, we sample from preliminary
attention windows. Pixels inside the windows are labeled with the
positive class. As to the negative class, since the pixel number outside
the window is larger than that of inside the window, we apply random
sampling to the outside region to balance the positive and negative
sample numbers. Furthermore, we discard pixels along the boundaries of
the attention windows within a certain distance threshold (say, 3
pixels) since their features are similar with each other. These pixels
are treated as noisy samples and are not included in model training.

\begin{figure}[t]
\begin{center}
\includegraphics[width=0.9\linewidth]{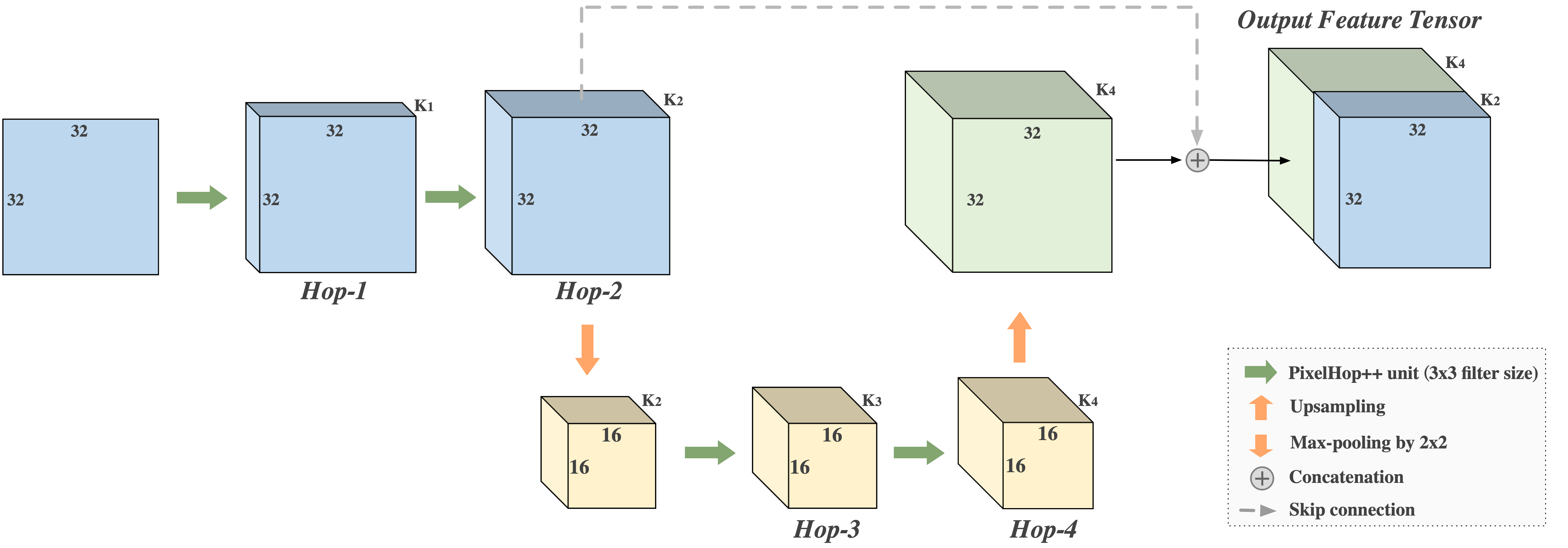}
\end{center}
\caption{Illustration of feature extraction in the attention map refinement step.}\label{fig:feat_att_2}
\end{figure}

In implementation, we use four cascaded PixelHop++ units to extract
features for the binary classification as shown in
Fig.~\ref{fig:feat_att_2}. Padding is used before each hop unit to keep
the resolution unchanged. To combine features from multiple scales for
the same spatial location, channel-wise upsampling by a factor of 2 is
applied to the Hop-4 feature map. Hop-4 features have a larger receptive
field. They are concatenated with Hop-2 features that have a smaller
receptive field through skip connection. The output feature tensor has a
good balance between global and local representations. 

\subsection{Step 3: Attention Region Finalization}\label{sec:rect_attention_final}

\begin{figure}[t]
\begin{center}
\includegraphics[width=0.8\linewidth]{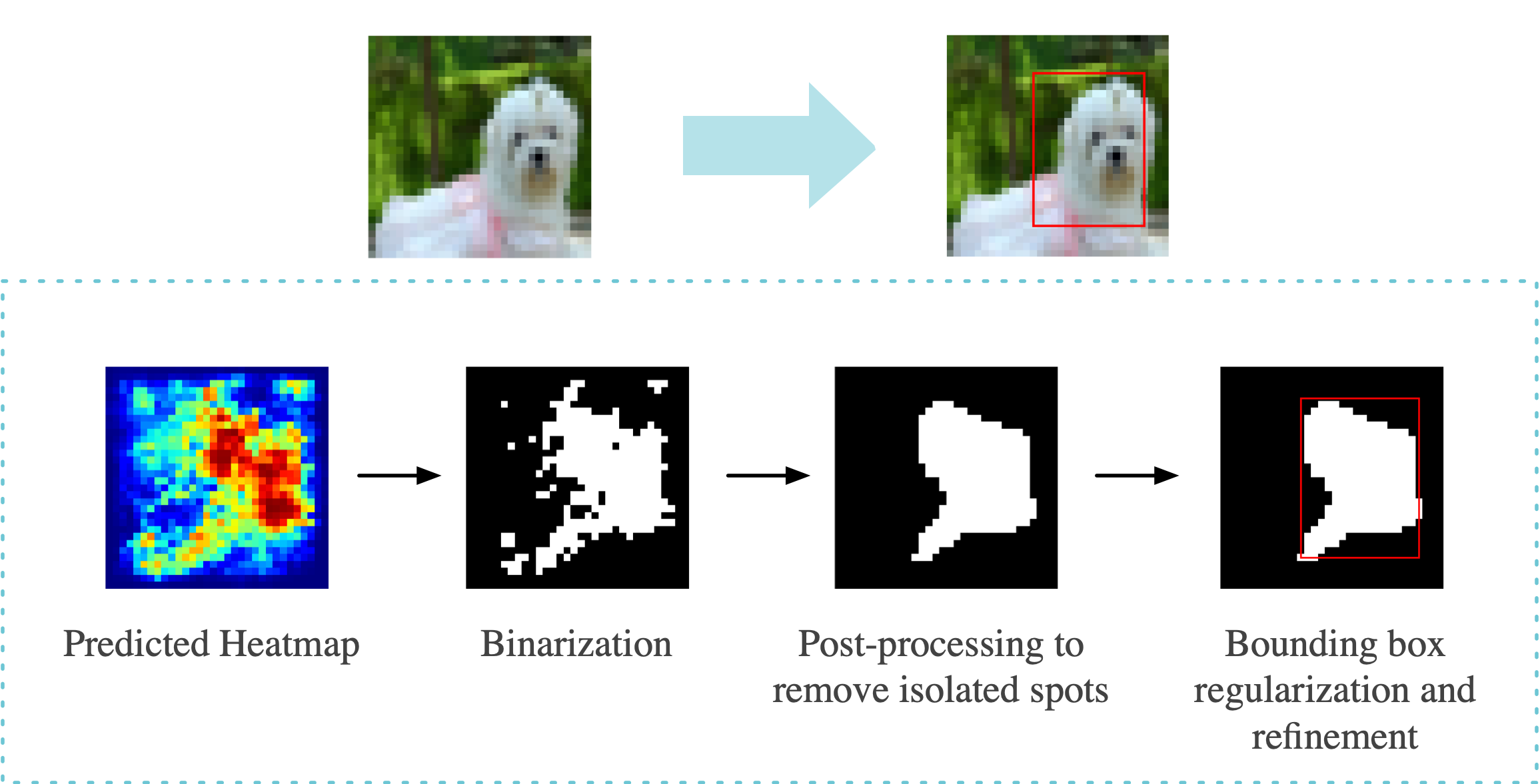}
\end{center}
\caption{Illustration of the attention region finalization step.}
\label{fig:bbox_pipe}
\end{figure}

Given the refined attention map from Step 2, we attempt to obtain a
rectangular attention region that is adaptive to various object sizes
and shapes in Step 3.  As illustrated in Fig. \ref{fig:bbox_pipe}, its
processing consists of the following three substeps.
\begin{itemize}
\item[a] Attention map binarization and cleaning: \\
The soft decision scores in the attention map represent the probability
of each pixel to be a foreground pixel. We set a uniform threshold
$T_{att}$ to binarize the decision values.  Empirically, we use
$T_{att}=0.5$. To remove the isolated spots, we apply a median filter
with radius 3 to each pixel. 

\item[b] Bounding box regularization: \\
The attention region is a rectangular one. We apply the
maximum-occupancy-rate pooling strategy to derive the tightest bounding
box that includes foreground pixels after the above substep.  The
estimated bounding box is regularized with its center location and its
height $H$ and width $W$. To avoid extremely small bounding boxes, if
$\max(H,W)$ is smaller than a threshold, the bounding box is enlarged to
$\max(H,W)=16$ while keeping the aspect ratio unchanged. 


\item[c] Final attention region extraction and resizing: \\
Based on the regularized bounding box, the corresponding region is
cropped out from the whole image. It is resized using the Lanczos
interpolation to the same resolution as the original input image, for
example, 32x32 for the CIFAR-10 dataset. 

\end{itemize}

\subsection{SAL-Assisted E-PixelHop}\label{sec:overall}

\begin{figure*}[t]
\begin{center}
\includegraphics[width=1.0\linewidth]{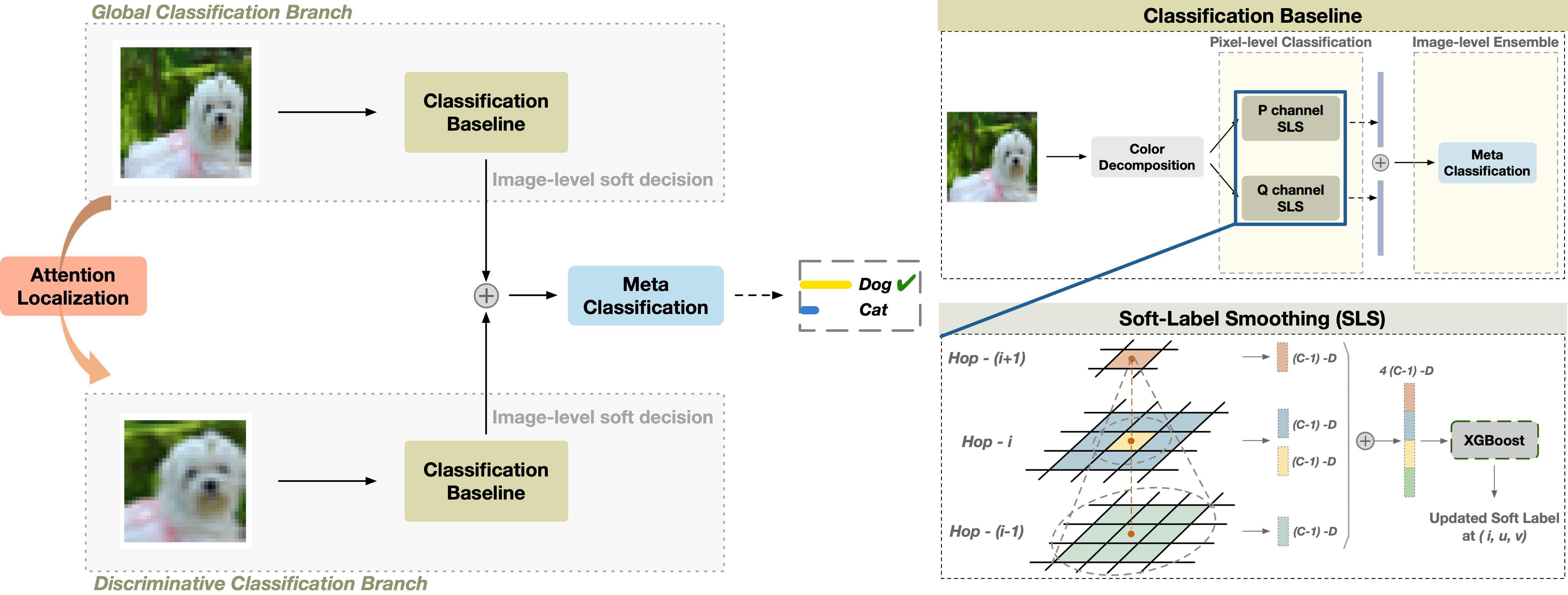}
\end{center}
\vspace{-5pt}
\caption{An overview of SAL-assisted E-PixelHop.}\label{fig:class_pipe}
\end{figure*}

As an application of SAL, we propose an object classification pipeline
with attention localization, named SAL-assisted E-PixelHop.  It contains
a two-stage decision pipeline: 1)
multi-class classification and 2) binary-class classfication among the
top two contenders. 

In Stage 1, we adopt E-PixelHop~\cite{yang2021pixelhop} as the baseline (see Fig. \ref{fig:class_pipe} right subfigure),
which is an SSL-based object classification solution. The color
representation is first converted from the RGB space to the PQR
space~\cite{yang2021pixelhop}.  Soft-label smoothing (SLS)~\cite{yang2021pixelhop} is performed
using cross-hop label update for reliable local decisions. Pixel
classification is conducted based on smoothed soft labels. Then,
pixel-wise decisions are flattened and ensembled through a meta
classifier for image label prediction. However, there is one minor difference
in this work. That is, the classification model is trained for a second
round on hard samples to further boost the performance of E-PixelHop. 

Fig. \ref{fig:class_pipe} illustrates the pipeline of Stage 2. It is
only applied to the top 2 contenders after Stage 1, which is called the confusing
set. For 10 object classes, there are at most 45 confusion sets.  Stage
2 has two branches: the global classification and the attention
classification. The former takes the whole image as its input while the
latter takes the cropped-out and resized attention region obtained from
SAL as input.  Furthermore, classification results of both branches
are ensembled through a logistic regression, which offers the highest
classification accuracy. 

\section{Experiments}\label{sec:experiments}

We visualize results obtained by SAL and conduct the performance of the
attention-based object classification task in this section. Our
experiments are carried out on the CIFAR-10~\cite{cifar10} dataset,
which contains 10 classes of tiny color images of spatial resolution
$32\times 32$ with 50,000 training and 10,000 test images. 

\begin{figure}[t]
\begin{center}
\includegraphics[width=0.8\linewidth]{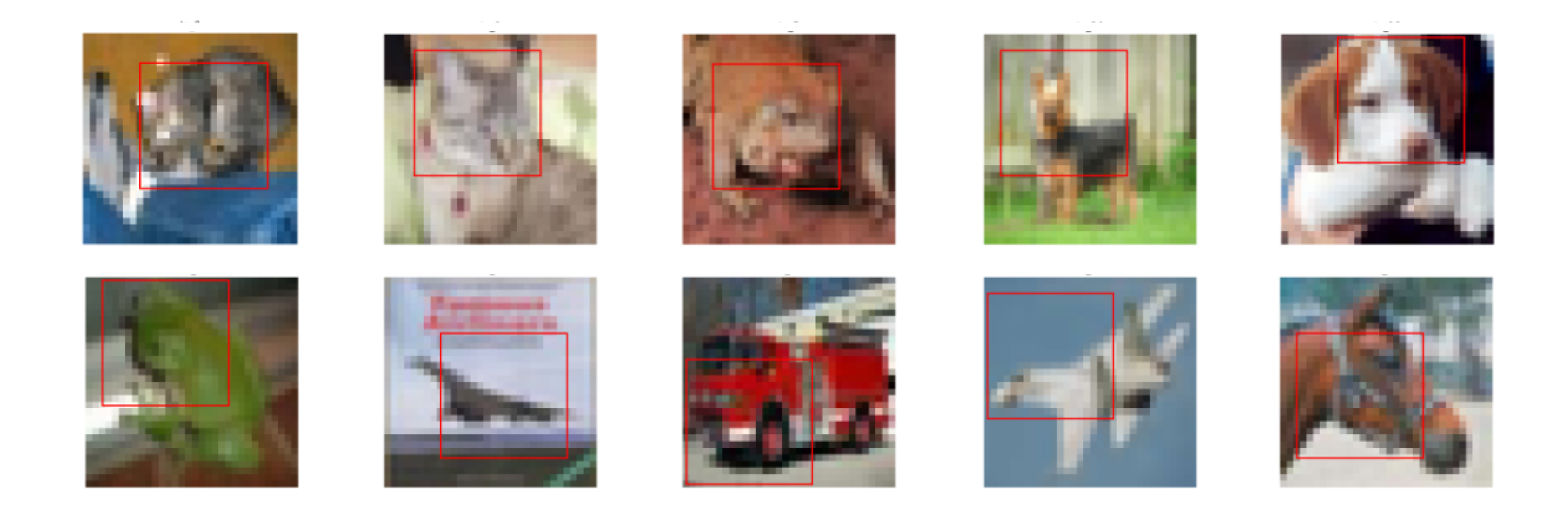}
\end{center}
\vspace{-15pt}
\caption{Results of preliminary attention windows for test 
images as indicated by red bounding boxes.}\label{fig:att_example1}
\end{figure}

\begin{figure}[htbp]
\begin{center}
\includegraphics[width=0.7\linewidth]{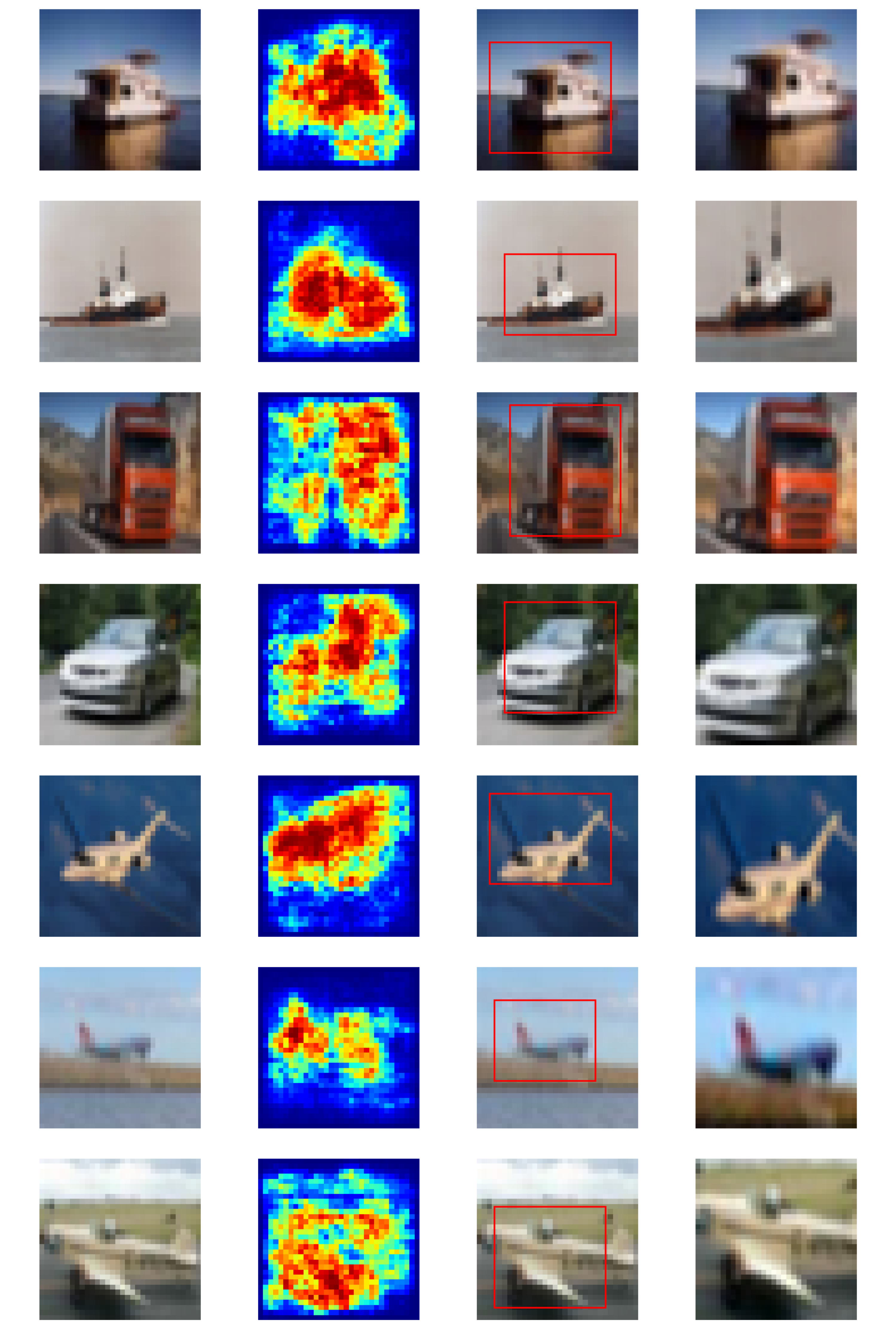}
\end{center}
\caption{Visualization of SAL results for non-animal images (from
left to right): the input image, the attention map (after Step 2 of SAL), 
the finalized attention region, and the cropped-out and rescaled image.} 
\label{fig:att_result_1}
\end{figure}

\begin{figure}[htbp]
\begin{center}
\includegraphics[width=0.7\linewidth]{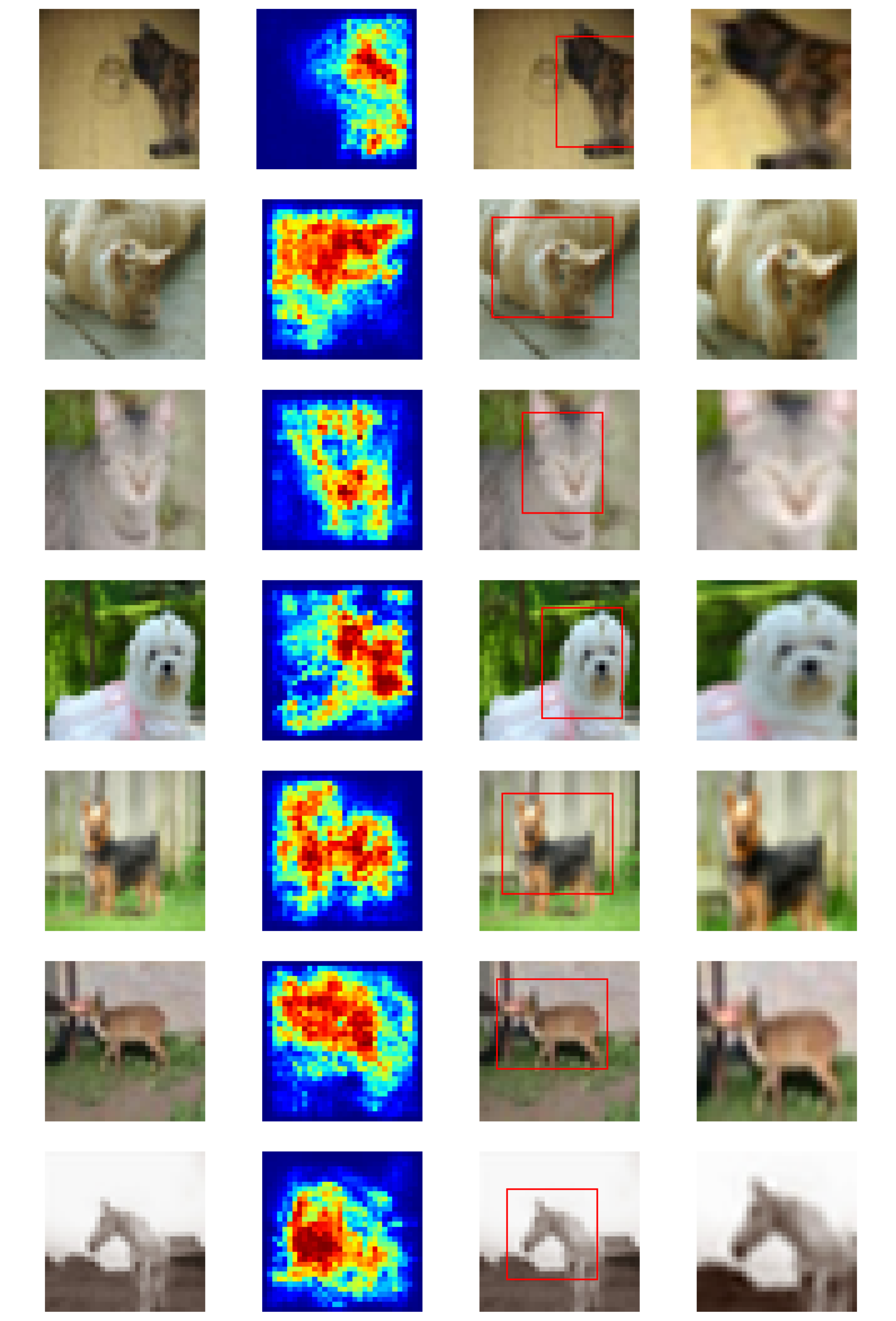}
\end{center}
\caption{Visualization of SAL results for animal images (from left to
right): the input image, the attention map (after Step 2 of SAL), the
finalized attention region, and the cropped-out and rescaled image.}
\label{fig:att_result_2}
\end{figure}

{\bf Visualization of SAL Results.} We show the preliminary attention
extraction results (i.e., after Step 1 of SAL) of 10 test images in
Fig.~\ref{fig:att_example1}, where the red bounding boxes indicate a
squared attention window of size $19 \times 19$. We see that it can
localize salient regions of an object quite well even without Steps 2
and 3.  However, the bounding box of a fixed size may not fit the object
tightly for some images (e.g., the rightmost image in the second row.)
Figs.~\ref{fig:att_result_1} and \ref{fig:att_result_2} show examples
after attention refinement and rectangular attention region finalization
for non-animal and animal objects, respectively.  Each row has four
subfigures. From left to right, they are the input image, the attention
map (i.e., after Step 2 of SAL), the finalized attention region, and
the cropped-out and resized attention image. It is shown qualitatively
in these examples that SAL can capture salient regions of low resolution
images well. It is robust to both large and small objects of various
rectangular shapes. 

{\bf Two-Class Object Classification.} Among the $10$ object classes in
CIFAR-10, the four most challenging pairs are Cat/Dog, Airplane/Ship,
Automobile/Truck, and Deer/Horse. There are 10K training images and 2K
test images in each pair. The challenges come from high similarity in
object shape (e.g.  the pose and the natural body outline between cats
and dogs, or deers and horses) as well as color tone of foreground or
background (e.g.  the blue sky for airplanes and blue ocean for ships).
We focus on these four pairs to evaluate the performance of SAL-assisted
binary object classification. 

We show the performance of two-class object classification in Table
\ref{tab:results_2k}. It gives the test accuracy for each pair under
three scenarios: (1) classification based on original images only; (2)
classification based on attention regions only; (3) ensemble results of
(1) and (2). On one hand, we see that results with attention regions
alone are not as competitive as those of the whole images in three out
of four cases.  This is because that CIFAR-10 images are tiny
low-resolution images, where the full images already focus on objects.
The contribution of attention may not be that visible. On the other
hand, one can see that the ensemble results outperform the best of the first
two.  Thus, it means that SAL provides meaningful auxiliary information
for higher classification accuracy.  

\begin{figure}[h]
\begin{center}
\includegraphics[width=0.6\linewidth]{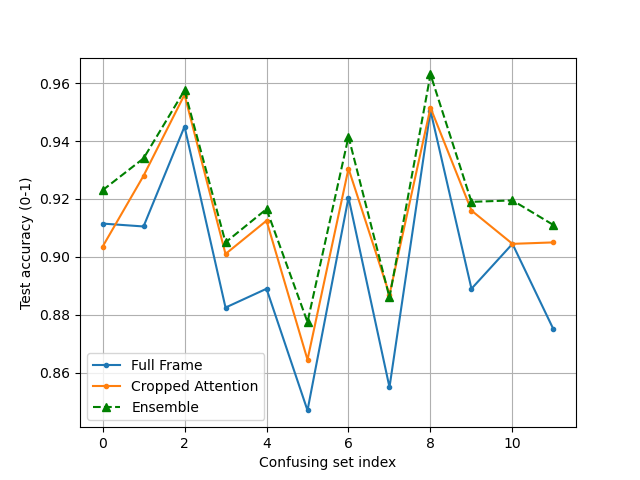}
\end{center}
\caption{Comparison of classification accuracy of test images for
another 11 confusion pairs among the top 15 confusing ones, where the 4
pairs reported in Table~\ref{tab:results_2k} are excluded.}
\label{fig:curve_2K_moregroup}
\end{figure}

Furthermore, we compare another eleven pairs among fifteen most
confusing pairs in Fig. \ref{fig:curve_2K_moregroup}. We see from the
figure that the attention region obtained by SAL yields better
classification results than those based on the full images in ten (out
of eleven) confusing pairs.  This provides another evidence of the power of SAL.
Again, the ensemble results perform the best. 

\begin{table}[b]
\small
\centering
\caption{Comparison of image-level test accuracy (\%) of the binary
object classification with and without attention localization}
\label{tab:results_2k}
\begin{tabular}{@{}lccc@{}} 
\toprule
                    & Full Frame    & Cropped Attention     & Ensemble          \\ 
\midrule
Cat vs Dog          & 79.10         & 77.90                 & \textbf{80.05}    \\
Airplane vs Ship    & 93.75         & 92.20                 & \textbf{94.10}    \\
Automobile vs Truck & 92.95         & 93.45                 & \textbf{93.90}    \\
Deer vs Horse       & 90.95         & 90.30                 & \textbf{92.30}    \\ 
\bottomrule
\end{tabular}
\end{table}

{\bf Ten-Class Object Classification.} For a 10-class classification
problem, we consider a two-stage decision pipeline. For the first-stage
classification, we conduct a baseline 10-class object classification
task and assign each test a soft-decision vector of 10 dimensions, where
each dimension indicates the probabilities of belonging to one class.
Images with the same top-2 candidate classes in the baseline decision
form one confusion set.  There are at most 45 confusion sets for
one-versus-one competition. The image number in each confusion set
varies.  For the second-stage classification, we focus on confusion sets
that have a large number of images and conduct the two-class
classification task. The class with a higher probability is chosen to
be the final decision.

The ablation study of adopting various strategies is summarized in Table
\ref{tab:ablation}. Our study focuses on the stage-2, which includes the
selected number of resolved confusion sets (25 or 45), the use of full
images only, the use of attention regions only, the ensemble of the two.
The final test accuracy for CIFAR-10 is given in the last column.  When
resolving all 45 confusion sets, the accuracy of stage-2 with ensembles
can outperform that without SAL at all by 1.06\%.

\begin{table}[t]
\small
\centering
\caption{Ablation study of 10-class classification accuracy with
different decision strategies for CIFAR-10.} \label{tab:ablation}
\begin{tabular}{@{}cccccc@{}}
\toprule
\multirow{3}{*}{Stage-1} & \multicolumn{4}{c}{Stage-2} & \multirow{3}{*}{Test Accuracy} \\ \cline{2-5}
                         & \# of Resolved   & Full      & Cropped       & \multirow{2}{*}{Ensemble} &        \\
                         & Confusing Sets   & Frame     & Attention     &       &                \\ 
\midrule
\checkmark &                &            &       &      & 76.54      \\[6pt]
\checkmark & 25             & \checkmark &       &       & 77.62          \\
\checkmark & 25             &       & \checkmark &       & 77.36          \\
\checkmark & 25             &       &       & \checkmark & 78.60           \\[6pt]
\checkmark & 45             & \checkmark &       &       & 77.72          \\
\checkmark & 45             &       & \checkmark &       & 77.49          \\
\checkmark & 45             &       &       & \checkmark & \textbf{78.78} \\
\bottomrule
\end{tabular}
\end{table}

Finally, we compare the performance of modified LeNet-5
~\cite{kuo2019interpretable} and four SSL-based classification systems
on CIFAR-10 in Table~\ref{tab:benchmark_2}. The four SSL-based solutions
include PixelHop \cite{chen2020pixelhop}, PixelHop$^+$
\cite{chen2020pixelhop}, PixelHop++ \cite{chen2020pixelhop++}, and
E-PixelHop~\cite{yang2021pixelhop}.  The two proposed SAL-assisted
E-PixelHop methods outperform all other benchmarking methods. 

\begin{table}[htbp]
\small
\centering
\caption{Comparison of testing accuracy (\%) of LeNet-5, PixelHop, 
PixelHop$^+$, PixelHop++, E-PixelHop and the proposed SAL-assisted E-PixelHop on CIFAR-10.}\label{tab:benchmark_2}
\begin{tabular}{@{}lc@{}} 
\toprule
                                        & Test Accuracy (\%) \\ 
\midrule
Lenet-5                                 & 68.72              \\[6pt]
PixelHop \cite{chen2020pixelhop}        & 71.37              \\
PixelHop$^+$ \cite{chen2020pixelhop}    & 72.66              \\
PixelHop++ \cite{chen2020pixelhop++}    & 66.81              \\ 
E-PixelHop \cite{yang2021pixelhop}      & 76.18             \\[6pt]
Ours - 25 Resolved Sets                 & \underline{78.60}            \\ 
Ours - 45 Resolved Sets                 & \textbf{78.78}            \\
\bottomrule
\end{tabular}
\end{table}

\section{Conclusion and Future Work}\label{sec:conclusion}

An SAL method to extract a distinctive object region was proposed in
this work. It was shown by experiments that SAL can localize attention
regions well for objects of various sizes and shapes in low resolution
images. SAL was used to enhance the performance of an object
classification baseline, E-PixelHop. The SAL-assisted E-PixelHop
outperforms all existing SSL-based classification systems for the
CIFAR-10 dataset. In the future, we plan to apply SAL to object images
of higher resolution such as ImageNet. 

\bibliographystyle{ieeetr}
\bibliography{refs}

\end{document}